\documentclass{article} % For LaTeX2e
\usepackage{iclr2023_conference,times}

\usepackage{amsmath,amsfonts,bm}

\def\figref#1{figure~\ref{#1}}
\def\secref#1{section~\ref{#1}}
\def\eqref#1{equation~\ref{#1}}
\def\1{\bm{1}}

\DeclareMathAlphabet{\mathsfit}{\encodingdefault}{\sfdefault}{m}{sl}
\SetMathAlphabet{\mathsfit}{bold}{\encodingdefault}{\sfdefault}{bx}{n}

\definecolor{citecolor}{HTML}{0071bc}
\usepackage[pagebackref=false,breaklinks=true,colorlinks,citecolor=citecolor,bookmarks=false]{hyperref}
\usepackage{url}

\usepackage{booktabs}       % professional-quality tables
\usepackage{xcolor}         % colors
\usepackage{soul} 

\usepackage{subfigure}
\usepackage{color}
\usepackage{multirow}
\usepackage{overpic}
\usepackage{wrapfig}
\usepackage{bbding}

\usepackage{silence}
\newcommand{\gsh}[1]{{\textcolor{black}{#1}}}

\newcommand{\myPara}[1]{\vspace{-.05in} \noindent\textbf{#1}}

\def\ie{\emph{i.e.,~}}
\def\eg{\emph{e.g.,~}}

\def\Real{\mathbb{R}}
\renewcommand{\figref}[1]{Fig.~\ref{#1}}%
\newcommand{\tabref}[1]{Tab.~\ref{#1}}%
\renewcommand{\secref}[1]{Section~\ref{#1}}
\renewcommand{\eqref}[1]{Equ.~(\ref{#1})}

\title{Towards Sustainable Self-supervised Learning}

\author{Shanghua Gao$^{1,2}$\thanks{This work was done while Shanghua Gao was a research intern at Sea AI Lab.} \quad
Pan Zhou$^{1}$ \quad
Ming-Ming Cheng$^{2}$ \quad
Shuicheng Yan$^{1}$ \\
$^1$Sea AI Lab \quad $^2$Nankai University \\
\texttt{\{gaoshanghua,zhoupan,shuicheng.yan\}@sea.com, cmm@nankai.edu.cn} \\
}

\iclrfinalcopy % Uncomment for camera-ready version, but NOT for submission.
\begin{document}

\maketitle
\begin{abstract}

Although increasingly training-expensive, 
most self-supervised learning (SSL) models 
have repeatedly been trained from scratch
but not fully utilized, since 
only a few SOTAs are employed for downstream tasks. 
In this work, we explore a sustainable SSL framework 
with  two major challenges: 
i) learning a stronger new SSL model based on the existing 
pretrained SSL model, also called as `` base" model,  in a cost-friendly manner,
ii) allowing the training of the new model to be compatible with
various base models. 
We propose a Target-Enhanced Conditional (TEC) scheme 
which introduces two components to the existing mask-reconstruction based SSL.  
Firstly, we propose patch-relation enhanced targets which enhances the target given by base model and  encourages 
the new model to learn semantic-relation knowledge from the base model by using   
incomplete inputs. 
This hardening and target-enhancing help the new model surpass the base model, 
since they enforce additional patch relation modeling to handle incomplete input.  
Secondly, we introduce a conditional adapter that 
adaptively adjusts new model prediction to align with the target of different base models. 
Extensive experimental results show  that our TEC scheme can accelerate the 
learning speed, and also improve SOTA SSL base models, \eg MAE and iBOT,  
taking an explorative step towards sustainable SSL.
    
\end{abstract}

\section{Introduction}
\vspace{-10pt}
\label{sec:intro}
Self-supervised learning (SSL) has achieved overwhelming success 
in unsupervised representation learning, 
with astonishingly high performance in many downstream tasks 
like classification~\citep{zhou2021ibot,mugs2022SSL}, 
object detection, and segmentation~\citep{bao2021beit,he2022masked}.  
In SSL, a pretext task is first built, 
\eg instance discrimination task~\citep{He_2020_CVPR,chen2021mocov3} 
or masked image modeling (MIM)~\citep{bao2021beit,he2022masked},   
and then pseudo labels are generated via the pretext task to 
train a network model without requiring manual labels.  
Though successful, SSL is developing towards a direction of 
requiring increasingly large training costs,  
\eg 200 training epochs in MoCo~\citep{He_2020_CVPR}  
while 16,00 epochs in MAE~\citep{he2022masked} to release its potential. 
Unfortunately,  most researchers only have limited computational budgets 
and often cannot  afford to train large SSL models. 
Moreover, the pretrained non-SOTA SSL models are rarely used in practice, 
since  SOTA is updated frequently and a previous one quickly becomes useless, 
wasting huge training resources.  
Thus, a \textbf{sustainable SSL} framework is much demanded. 

\begin{wrapfigure}{r}{6cm}
	\label{fig:sus}
	\centering
	\vspace{-10pt}
	\scriptsize
	\begin{overpic}[width=\linewidth]{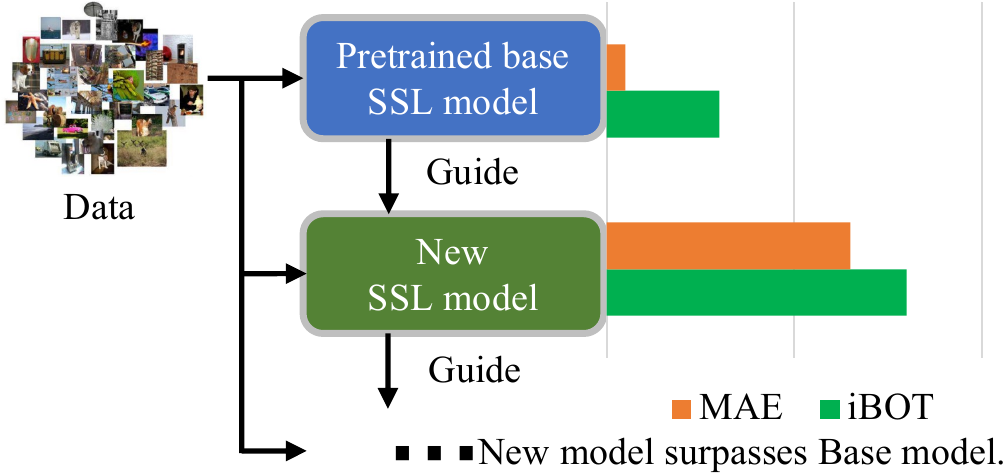} %
		\put(62.5,39){83.6\%}
		\put(72,33.7){84.1\%}
		\put(85,21){84.8\%}
		\put(90,16.4){85.1\%}
	\end{overpic}
	\vspace{-20pt}
	\caption{The concept of sustainable SSL.}
	\vspace{-10pt}
\end{wrapfigure}
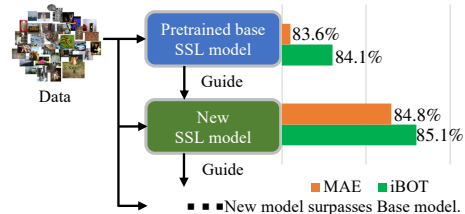

Just like how human experience is enriched and passed from one generation 
to the next in human society, 
we try to let an SSL model inherit the knowledge from a pretrained SSL base model 
to achieve superior representation learning ability for  %reusable or
 ``sustainable" learning and also to improve  learning efficiency than training a new SSL model from scratch.   
\figref{fig:sus}~illustrates the sustainable SSL for more clarity, in which we call the new SSL model to be trained as the new model 
and the pretrained SSL model as the base model.  %
To surpass the base model, in sustainable SSL, 
the new model exploits not only the implicit base model knowledge 
but also the absent knowledge in the base model.
Such a learning process follows a fully self-supervised manner
and differs from the self-training schemes
\citep{xie2020self,yalniz2019billion}  
that require labels for supervised learning.  This process is somewhat like Knowledge Distillation (KD)
\citep{hinton2015distilling,gou2021knowledge}, but differs from KD essentially.
For  KD, it aims to compress knowledge from a powerful teacher model 
to a compact student model which often suffers from performance degradation, while  our  sustainable SSL   targets at learning a more powerful new model based on  the base model.  

\newcommand{\addImg}[1]{\includegraphics[width=0.175\linewidth]{figures/imgs/img/#1.jpg}}
\newcommand{\addtec}[1]{\includegraphics[width=0.175\linewidth]{figures/imgs/tec/#1.jpg}}
\newcommand{\addibot}[1]{\includegraphics[width=0.175\linewidth]{figures/imgs/ibot/#1.jpg}}

\begin{wrapfigure}{r}{5.8cm}
	\vspace{-10pt}
	\centering
	\renewcommand{\arraystretch}{0.2}
	\setlength{\tabcolsep}{0.2mm}
	\begin{tabular}{lcccccccc}
		\rotatebox{90}{ IMG} &\addImg{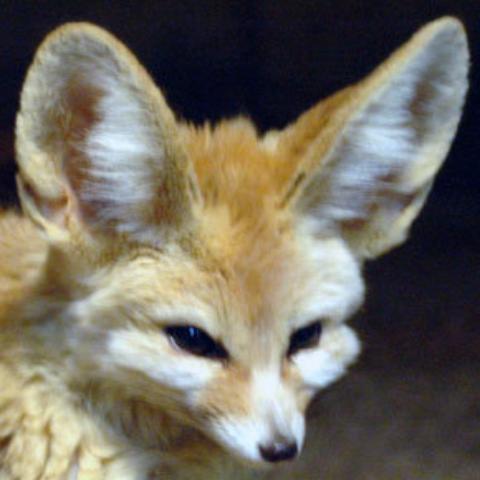} & \addImg{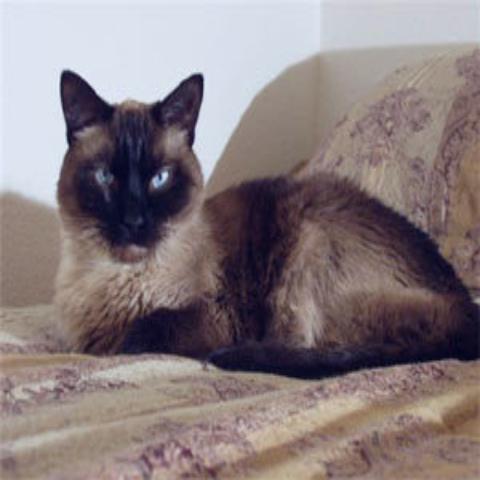}& \addImg{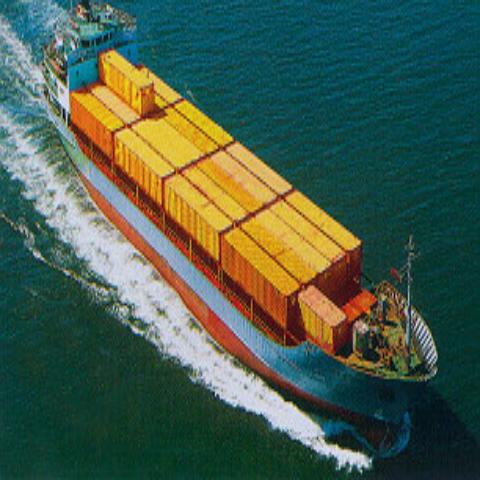}& \addImg{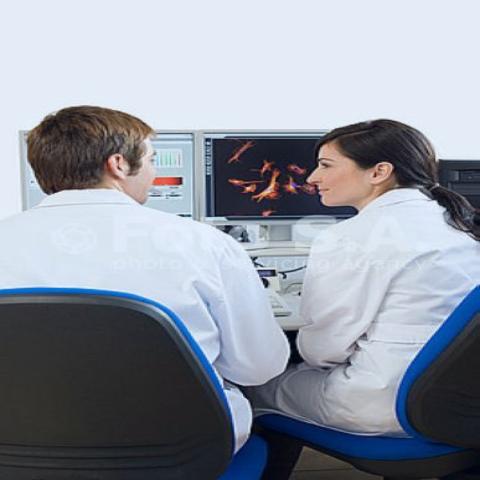}& \addImg{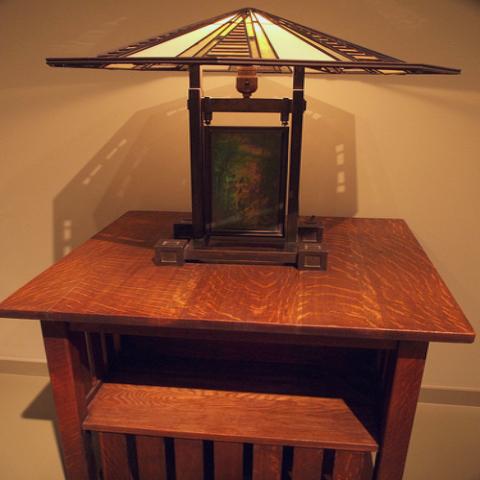} \\
		\rotatebox{90}{ iBOT}&\addibot{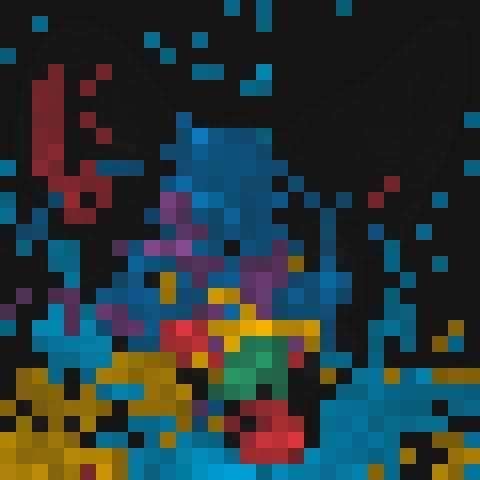} & \addibot{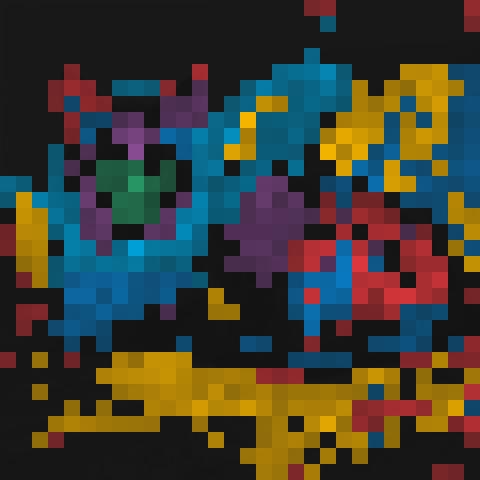}& \addibot{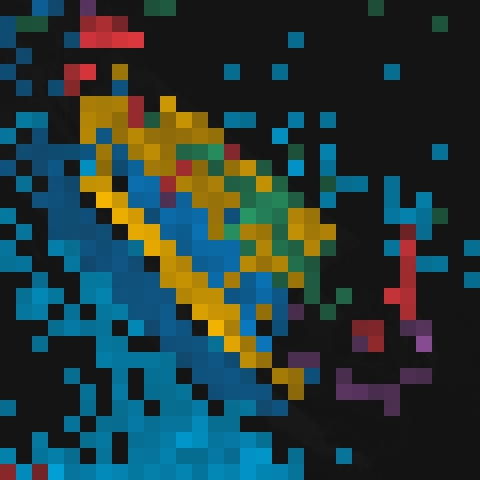}& \addibot{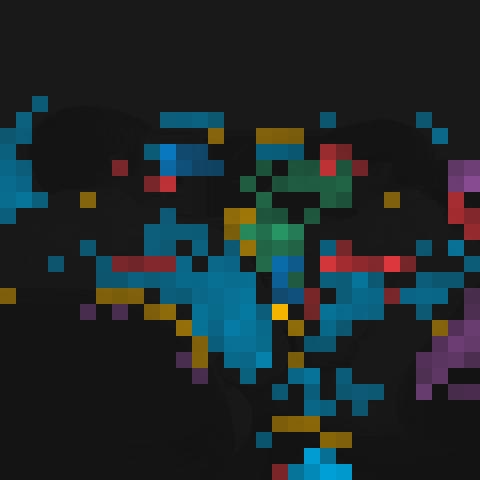}& \addibot{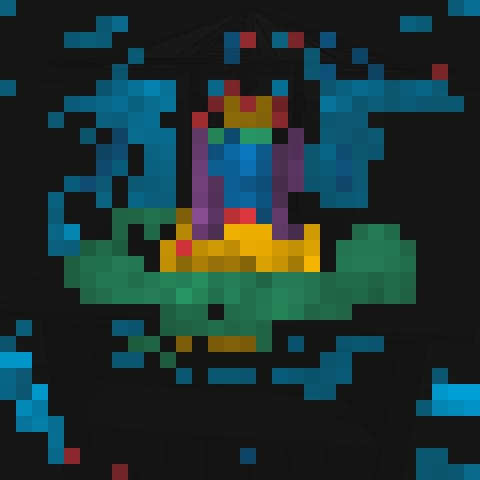} \\
		\rotatebox{90}{ TEC} &\addtec{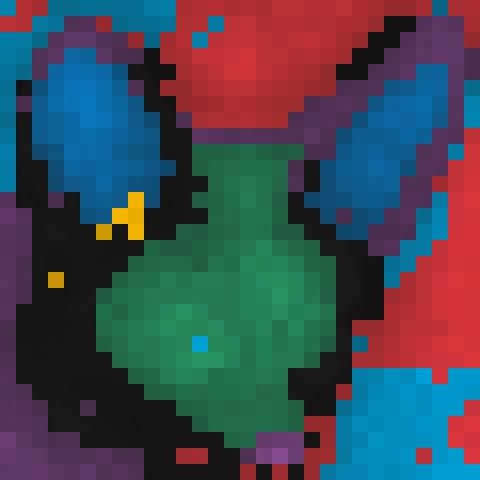} & \addtec{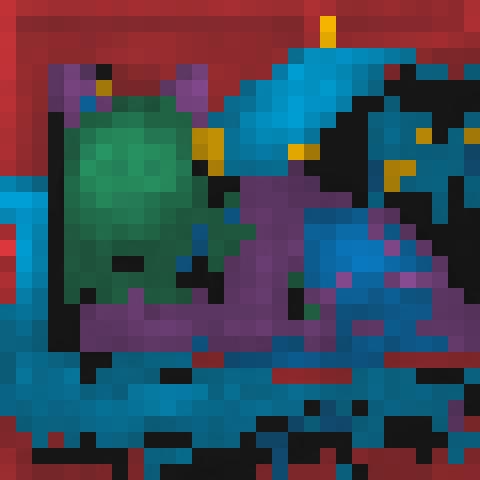}& \addtec{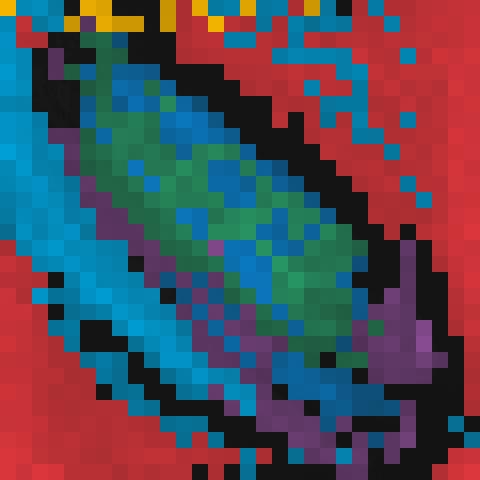}& \addtec{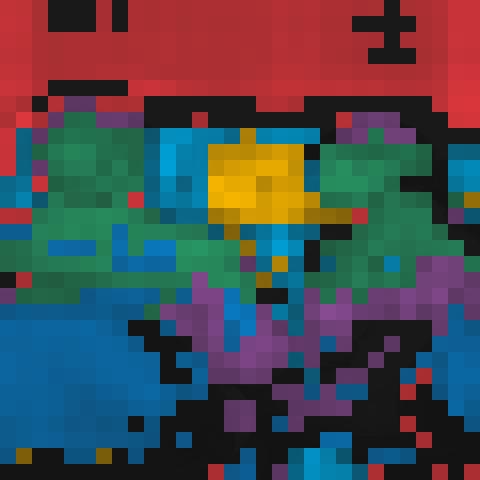}& \addtec{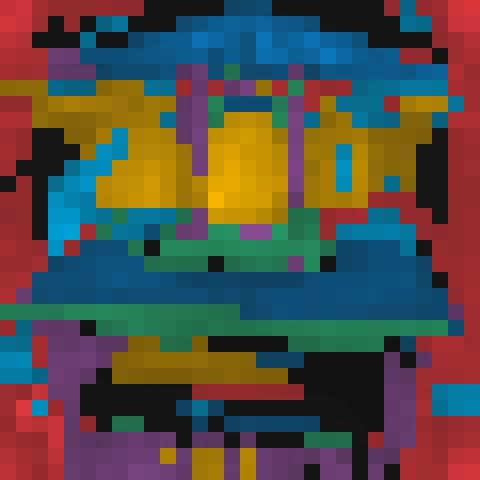} \\
	\end{tabular}
	\vspace{-9pt}
	\caption{Self-attention visualization.  
		Different colors denote attentions of different heads. 
		Black means no attention.
	}\label{fig:vis_att}
	\vspace{-8pt}
\end{wrapfigure}

In this work, we take an explorative step towards sustainable SSL 
by efficiently learning from existing pretrained SSL models and  surpassing them. 
In this work, to achieve this challenging goal, 
we encourage the new model to learn not only  knowledge of the base model but also   
more semantic-related new knowledge.  
We therefore choose a mask-reconstruction~\citep{he2022masked} SSL scheme to  
  train  the new model, 
in which the base model generates reconstruction targets 
from the full input images and the new model tries to predict the generated  
targets from randomly masked image input.
With this pretext task, the new model is forced to learn 
the semantics of the full input and its patch relations 
so that the new model can reason the desired full information from an incomplete input. 
As illustrated by~\figref{fig:vis_att}, 
the attentions of iBOT~\citep{zhou2021ibot} miss some semantic regions, \eg ears,  
while TEC with iBOT as the base model captures all semantics and 
well distinguishes all different components of an input image.  
Because of its more powerful ability to capture comprehensive semantic,  
TEC  helps achieve the challenging sustainable SSL, and actually can provide rich and flexible semantics for downstream tasks.

However, different SSL base models could have various properties due to 
their various training targets and training strategies, 
\eg iBOT models with more category semantics 
while MAE models with more image details~\citep{he2022masked}. 
So it is important to build high-qualified and compatible 
reconstruction targets from the base model so that 
the new model learns these targets in a complementary manner. 
A good model target should reveal the semantic relations among patches, 
\eg the relation between car wheels and car body, 
so that new model can learn these general relation patterns and 
adapts to downstream tasks.
To this end, we propose to enhance the target quality of the base model by using two complementary reconstruction targets: 
a) the patch-dim normalization which normalizes base model targets 
along patch dimension to enhance the relations among input patches, and 
b) patch attention maps with rich semantics to filter out 
possible noise and establish the correlation 
between the whole image semantic and the patch semantic.  
For target compatibility,  
we introduce conditional adapters into the new model so that new model predictions
can be adaptable to various base models with different properties. 
Given a base model target, 
adapters conditionally active and adjust mid-level features of the new model 
to predict the target more effectively. 
These adapters are discarded after pretraining
but can serve parameter-efficient finetuning
\citep{jia2022visual,chen2022adaptformer} if kept.

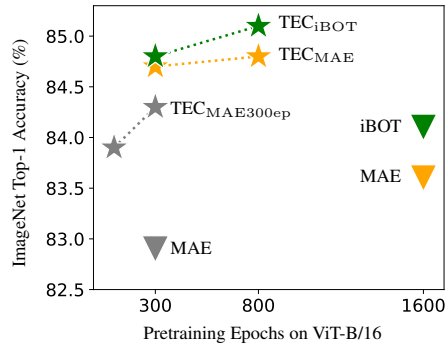
\begin{wrapfigure}{r}{6cm}
	\centering
	\vspace{-5pt}
	\scriptsize
	\begin{overpic}[width=\linewidth]{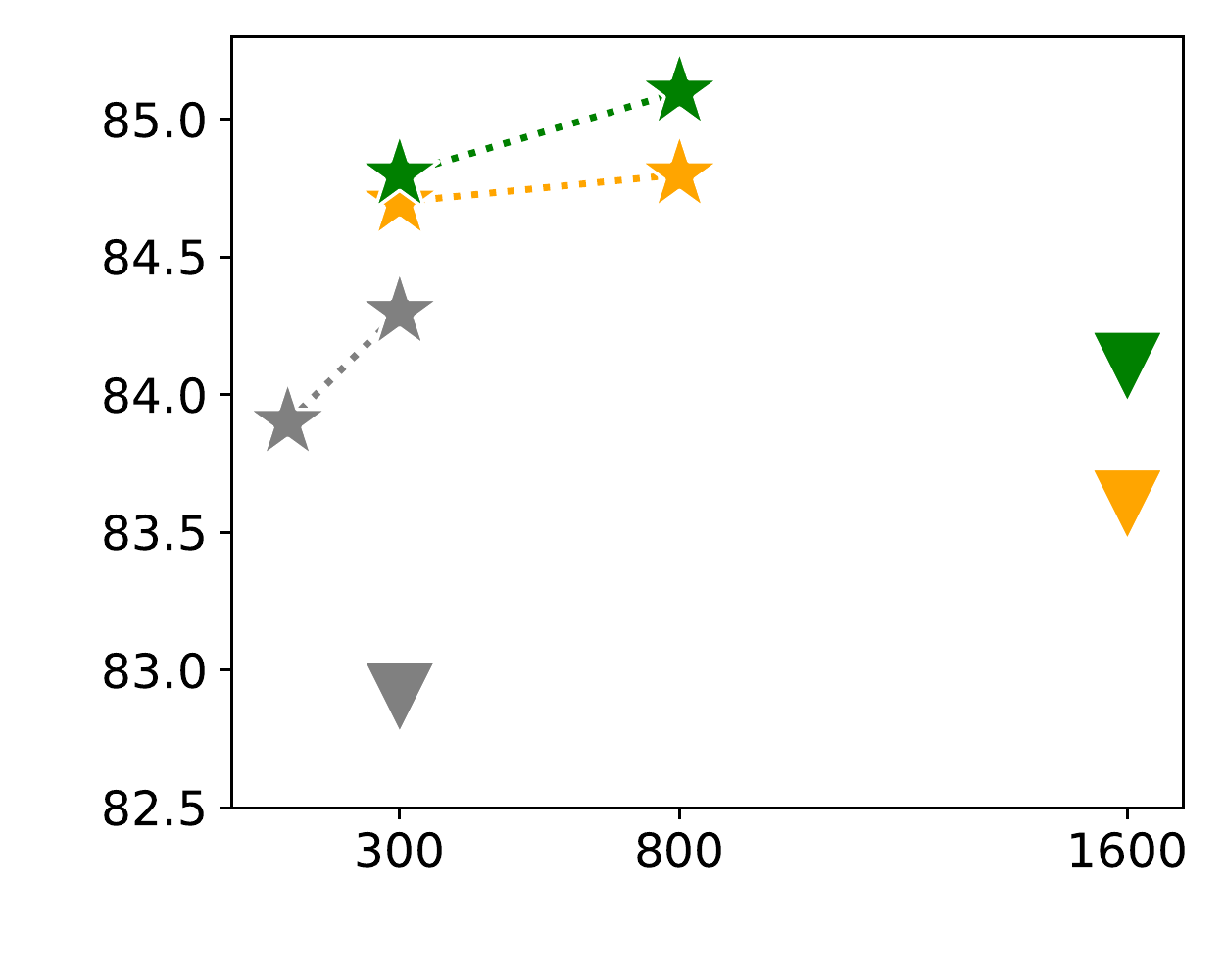} %
		\put(60,72){TEC$_{\rm iBOT}$}
		\put(60,65){TEC$_{\rm MAE}$}
		\put(36,53){TEC$_{\rm MAE300ep}$}
		\put(78,49){iBOT}
		\put(78,38){MAE}
		\put(36,22){MAE}
		\put(30,3){Pretraining Epochs on ViT-B/16}
		\put(1,19){\rotatebox{90}{ImageNet Top-1 Accuracy (\%)}}
	\end{overpic}
	\vspace{-21pt}
	\caption{Top1 accuracy on ImageNet-1k. TEC models have the same color with their base model.}
	\vspace{-10pt}
	\label{fig:epoch}
\end{wrapfigure}
We call the above method for sustainable SSL as 
Target-Enhanced Conditional (TEC) mask-reconstruction. 
As shown in~\figref{fig:epoch}, on ImageNet, 
TEC without any extra training data  
improves the SSL base model by a remarkable margin,  
\eg MAE~\citep{he2022masked} and iBOT~\citep{zhou2021ibot}. For instance, taking iBOT with 1600 epochs  as base model, TEC with only 800 training epochs makes 1.0\% improvement. 
Moreover, we also find that TEC can significantly accelerate 
the SSL learning process and saves training cost. 
For example, training TEC for only 100 epochs with random initialization 
and a 300-epochs-trained MAE base model outperforms MAE trained with 1600 epochs.
We hope our initial effort towards sustainable SSL will inspire more works in the future 
to sustainably improve SSL in a cost-friendly manner.

\section{Method}

\subsection{Overall framework}

\begin{figure}[t]
	\centering
	\tiny
	\begin{overpic}[width=\linewidth]{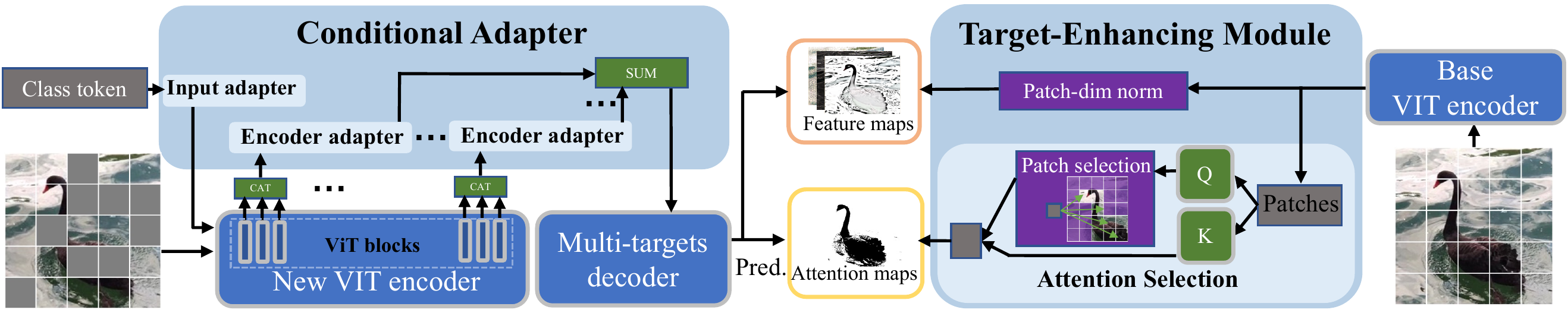} %
		\put(43,7.2){$Z_{e}$}
		\put(64.7,5){$A_{c}^{'}$}
		\put(73.5,4.2){$Z_{k}$}
		\put(73.5,7){$Z_{q}$}
		\put(61.5,8){$Z_{q}^{'}$}
		\put(60.5,4.1){$A_{s}$}
		\put(61,15){$Y_{f}$}
		\put(48,15){$Z_{f}$}
		\put(48,5.5){$Z_{a}$}
	\end{overpic} 
	\vspace{-10pt}
	\caption{The overall framework of the proposed TEC. 	
	  The pretrained SSL base model in TEC 
		generates patch-relation enhanced reconstruction targets, 
		\ie patch-dim normalized features and semantic attention maps.
	  The new ViT encoder takes in masked image and 
		the class token enhanced by the input adapter,   
		and then sequentially passes the generated features into encoder adapters 
		and the multi-target decoder to predict the targets given  by base model.
	}\vspace{-10pt}
	\label{fig:overall}
\end{figure}
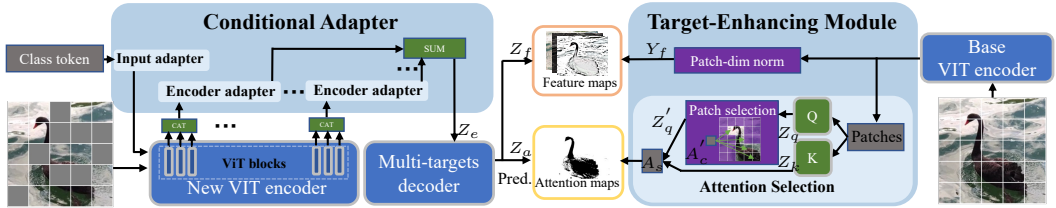

An overall framework of the proposed target-enhanced conditional (TEC) 
mask-reconstruction method is illustrated in~\figref{fig:overall}. 
TEC follows~\citep{he2022masked,bao2021beit}
to use Vision Transformer (ViT)~\citep{dosovitskiy2020image} 
for implementation. 
Under the mask-reconstruction framework~\citep{he2022masked}, as shown in~\figref{fig:overall}, TEC consists of a new ViT encoder to be pretrained, conditional adapters for conditional pretraining, a multi-target decoder for reconstruction targets prediction,    
an SSL pretrained ViT encoder as the base model,
and a target-enhancing module to 
construct patch-relation enhanced reconstruction targets from the base model. 
Specifically, the base model is an SSL-pretrained ViT encoder (\eg in MAE~\citep{he2022masked}) and is used to generate the latent semantic of a full image. 
Then  target-enhancing module enhances  the latent semantic  to construct two complementary  reconstruction targets  as the supervision of  the new model.  
The new ViT encoder together with  adapters  takes in masked images,  and generates adapted latent semantics that are then fed into the multi-target  decoder to predict the base model targets. 
After pretraining, the new ViT encoder is kept for downstream tasks while other parts are removed. 
At below,  we will introduce the conditional pretraining aided by adapters in~\secref{sec:adapt} which   helps the new model effectively predict base model targets, and elaborate on 
 the target-enhancing module to generate high-qualified  base model  targets  in~\secref{sec:multiobj}.  
% build 

\subsection{Conditional Pretraining}
\label{sec:adapt}
As aforementioned, base models often have different properties, \eg~more global category semantic in iBOT while more local details  in MAE.   
So the prediction of the new model should be compatible with any given base model. 
To resolve a similar issue on vanilla image pixel reconstruction, 
the works~\citep{wang2022repre,dong2022ict,gao2022convmae} manually select certain  features from the mid-level layers of the encoder by trial and error to better align with the image pixel target.   
However, it is almost impossible to manually select features from certain fixed layers that are  compatible with different base models because of their possible different properties.  
So  to better predict base model targets,	
the new model should have conditional adaptation ability regarding
a given SSL base model.

Given a fixed pretrained model, the parameter-efficient fine-tuning scheme introduces   
trainable  extra modules with a small number of parameters 
into this pretrained model for  adapting it to  downstream tasks  in both vision~\citep{jia2022visual,chen2022adaptformer} and NLP~\citep{houlsby2019parameter,li2021prefix,liu2021gpt} domains.
For example,
the prompting scheme~\citep{li2021prefix,liu2021gpt,jia2022visual}
concatenates learnable input tokens, \eg class token, with patch tokens 
to activate certain semantic features of a fixed ViT model that are suitable for specific downstream tasks. 
Also,
inserting lightweight adapter modules (\eg MLP~\citep{houlsby2019parameter,chen2022adaptformer} and residual blocks~\citep{li2022exploring})
into a fixed model can modulate mid-level features of the model to predict features required by the downstream task.
Inspired by these parameter-efficient fine-tuning schemes,
we introduce the adaptation scheme into the pretraining stage 
to handle the diversities of base models
by equipping the new model with conditional adapters. 
Since our adapters are only used for pretraining and will be removed during finetuning,   they do not increase extra inference costs. 
Actually,  \tabref{tab:prompting} shows that keeping these adapters in the inference phase  enhances the parameter-efficient finetuning ability of the model. At below, we will introduce how to apply adapters, \ie input and encoder adapters,
into the new model encoder.

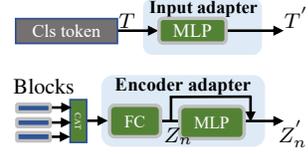
\begin{wrapfigure}{r}{4cm}
	\centering
	\vspace{-13pt}
	\setlength{\tabcolsep}{3.8pt} % column spacing
	\renewcommand{\arraystretch}{3.7}%  row spacing
	{ \fontsize{8.3}{1}\selectfont{
			\begin{overpic}[width=\linewidth]{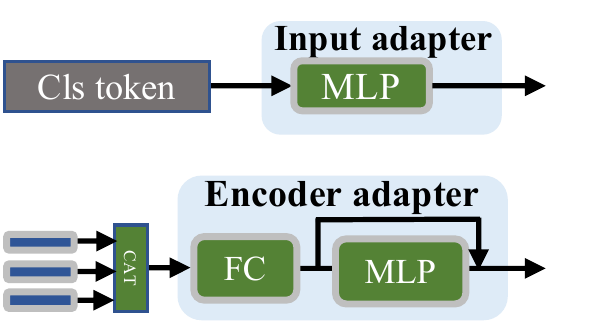} %
				\put(90,40){$T^{'}$}
				\put(35,40){$T$}
				\put(0,18){Blocks}
				\put(50,2){$Z_{n}$}
				\put(88,2){$Z_{n}^{'}$}
	\end{overpic} }}
	\vspace{-21pt}
	\caption{The input adapter and encoder adapter.}
	\vspace{-10pt}
	\label{fig:adapt}
\end{wrapfigure}
	\myPara{Input adapter.} 
For ViT networks, one often concatenates a class token  with the input patch tokens to learn the global semantics of the whole input. 
Since the prompting scheme shows the adaption ability of the class token, 
we propose to further enhance the feature adaption ability of the class token by adding an input adapter.  
As shown in \figref{fig:adapt},  the input adapter which is implemented by a small two-layer MLP layer enhances 
the representation ability of the class token 	so that the class token can better activate features in the new model 		according to the base model targets.
Specifically, the class token $T \in \Real^{C}$ of the ViT is processed by the MLP layer to obtain an enhanced class token $T^{'}  \in \Real^{C}$:
$$T^{'} = \mathbf{MLP}(T),$$ 
where $C$ is the embedding dimension. 
During pretraining, $T^{'}$ is appended to the patch tokens.
$\mathbf{MLP}$ enhances the representation ability of $T$ and enables the new model to better predict base model targets. 
For inference, since $\mathbf{MLP}(T)$ is shared by all input samples, 
one can compute it in advance to get $T^{'}$ as the new class token, 
meaning no extra cost is brought by $\mathbf{MLP}(T)$.

\myPara{Encoder adapter.}
To modulate mid-level features in the new model so that it can adapt to the base model targets, we apply a simple MLP with residual connection~\citep{chen2022adaptformer} as our encoder adapter in the  pretraining phase. 
As we hope to remove adapters after pretraining for higher inference efficiency,
we need to keep the encoder network topology unchanged after removing adapters. 
So we put the input of adapters in the middle of the encoder and merge all adapter outputs at the end of the encoder. 
As shown in~\figref{fig:adapt}, given features $X = \{X_{i}, i =  1, \ldots, D \}$ from each encoder block where $D$ is the encoder block number, we first uniformly divide them into $N$ groups, in which  each group contains 3 blocks by default. 
Within the $n_{th}$ group, we merge features from all blocks: 
\begin{equation*}
	Z_{n} = \mathbf{FC}(\mathbf{Concat}(X_i, ..., X_j)). 
\end{equation*}
Then we feed the feature $Z_{n}$ into an adapter and obtain an overall feature $Z_{e}$:  
\begin{equation}\label{targets}
	Z_{n}^{'}=Z_{n} + \mathbf{MLP}(Z_{n}), \qquad Z_{e} = \sum\nolimits_{n=1}^{N}{Z_{n}^{'}},
\end{equation}
where $\mathbf{MLP}$ is a small MLP of two fully-connected layers.	
The adapted features are then fed into the multi-target decoder to predict base model targets,  which will be introduced in~\secref{sec:multiobj}.

\vspace{-10pt}
\subsection{Patch-relation enhanced Reconstruction Targets}
\vspace{-5pt}
\label{sec:multiobj}
To better exploit the knowledge of base models for sustainable SSL, our target-enhancing module constructs two complementary targets with enhanced patch relations: 1) feature-level targets 
with patch dimension normalization to strengthen the relations among patches; 2) semantic 
attention maps to learn relations between semantic patches and other patches. The feature-level
targets reveal the semantics of certain patches, while attention maps focus more on  relations among patch feature.

\begin{wrapfigure}{r}{5cm}
	\vspace{-15pt}
   \centering
   \tiny
   \begin{overpic}[width=0.99\linewidth]{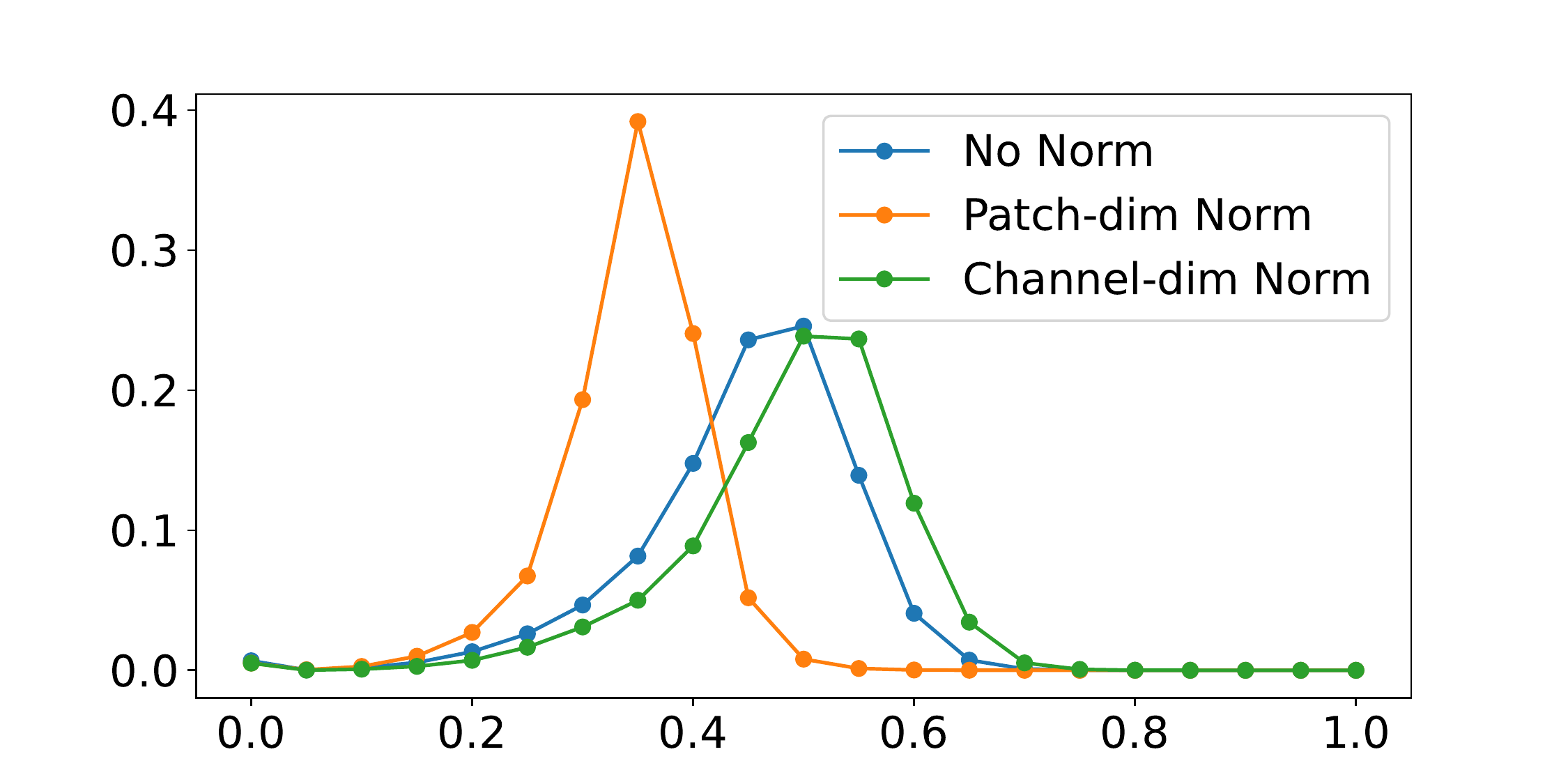} %
	   \put(3,20){\rotatebox{90}{Ratio}}
	   \put(33,-3){Patch Similarity}
   \end{overpic}
	   \vspace{-5pt}
	  \caption{The patch similarity distribution of MAE. } 
	 \label{fig:norm_compare}
	\vspace{2pt}
   \end{wrapfigure}
\myPara{Patch-dim normalized feature-level targets.} 
Given a base model, we  propose to normalize its target  
along the patch dimension to enhance the spatial patch relations.  
Specifically, for an input, assume its base model target is  $Y \in \Real^{L\times C}$ where  
$L$ and $C$ respectively denote patch number and  channel dimension. 
Then we normalize $Y$ along the patch dimension:  
\begin{equation}
	Y_{f} = (Y - \mu_{L})/\sigma_{L},
	\label{eq:norm}
\end{equation}
where $\mu_{L}$ and $\sigma_{L}$ are respectively mean and variance along the patch dimension.  For MIM, this patch-dim normalization can better enhance the spatial relations among tokens than the widely used feature normalization~\citep{wei2022contrastive,wei2022masked,baevski2022data2vec} along channel dimension. 
This is because as observed in Fig.~\ref{fig:norm_compare}, when using MAE for pretraining, the base model features of different patches actually have similar values and thus have high similarity score, since they all reveal the global semantic of the image. This cannot well reveal the  spatial relations among these patches. 
As a result,
the new model can easily reconstruct the feature target of masked
patches due to their high similarity with visible  patches, but does not well learn  spatial relations among patches.
Normalization along channel dimension can hardly enhance
the patch relations as it only consider the mean and variance within a patch. Actually, channel-dim normalization even enlarges the similarity among patches
as revealed in~\figref{fig:norm_compare}.
In contrast, patch-dim normalization ensures values within each channel has a clear difference 
and enhances the possible inherent spatial relations among patches by obviously reducing the similarity among patches as shown in~\figref{fig:norm_compare}. 
Moreover, \tabref{tab:norm_dim} shows that our normalization can significantly improves the new model  performance.   
After normalization, following~\citep{he2022masked}, the new model uses a fully-connected layer followed by the decoder to generate $Z_{f}$ for predicting the  base model target $Y_{f}$  on masked patches:
\begin{equation}
	L_{\rm fea} = \| M \circ (Y_{f} - Z_{f})\|_2^2,
	\label{eq:attfea}
\end{equation}
where $M$ is the mask matrix and $\circ$ denotes the element-wise product.

\myPara{Semantic attention-level targets.}
Self-attention  in pretrained ViT models has a powerful capability of capturing semantic relations among  patch tokens~\citep{caron2021emerging,li2022exploringself}.  
We then propose to utilize the self-attention map as a type of reconstruction target for MIM to further enhance the semantic relation modeling capability of the new model.
According to previous investigations on the effects of self-attention map in KD~\citep{wu2022self,wang2022closer}, not all attention map contains  useful semantic relations, and severe noisy attentions even hinder student learning. 
Accordingly, it is necessary to select parts of attention maps to reduce the possible severe noise and also help reduce training costs.

Here we utilize the base model class token that contains sufficient global semantics  
to select top similar patch tokens, which filters out the possible noises.  Specifically, given the attention maps  $A_{c} \in \Real^{H \times L}$ between class token and patch tokens  from the last ViT block in  base model where  $L$ and $H$ respectively denote patch number and head number,  
we  average  the attention map $A_{c}$  along head dimension to obtain $A_{c}^{'}\in \Real^{1 \times L}$. 
Then, as shown in~\figref{fig:overall}, we select top-$k$ patches with the largest values in $A_{c}^{'}$, and then compute the attention map $A_{p} \in \Real^{H \times k \times L}$ among the  top-$k$ patches and all patch tokens. Considering the importance of the class token, we further concatenate attention maps between itself and selected $A_{p}$ to obtain our final  reconstruction targets, \ie $A_{s} \in \Real^{H \times (k+1)\times L}$. 
Note, when we compute $A_s$, a temperature $\tau$ is added before the Softmax operation to adjust the attention sharpness.   
For the new model, we respectively use two fully-connected layers to project its decoder output into two predictions $Z_{q} \in \Real^{L \times C}$ and $Z_{k} \in \Real^{L \times C}$. 
We select the same patches as in $A_{s}$ from $Z_{q}$ to form $Z_{q}^{'} \in \Real^{(k+1) \times C}$. 
Then we   concatenate the class token $\textsf{cls}$ in new model  with  $Z_q'$ and compute  the KQ attention map  $Z_{a} = \textbf{\rm Softmax}([Z_{q}^{'}, \text{\textsf{cls}}]^{\top} Z_k)\in \Real^{H \times (k+1)\times L}$.  
Finally, we compute the predicted entropy  loss between the prediction  $Z_{a} $ and the   target  $A_s$  as 
\begin{equation}
	L_{\rm att} =  -A_{s} \log Z_{a}. 
\end{equation}

\myPara{Multi-target  decoder.}
Due to the different properties of two reconstruction targets,  namely feature target and attention target,  
one decoder in a new model for prediction is insufficient to handle them simultaneously and often   results in prediction  conflict. 
But  using separate decoders for each target would increase the trainable parameters and thus slow down the training.  To solve this issue, we use a simple decoder adaptation scheme that constructs target-specific inputs and then feeds them into a shared decoder.
Specifically, we feed the output feature $Z_{e}$
(see Eqn.~\ref{targets}) of the new model encoder into a fully-connected layer and then fill the masked tokens with a learnable mask token to obtain $Z_f^{'}$.  
Then similarly, given $Z_{e}$, we also use a fully-connected layer and a learnable mask token to obtain $Z_m^{'}$. 
Next, we respectively feed $Z_f^{'}$ and $Z_m^{'}$ into a shared transformer-based decoder for predicting the feature and attention map targets of the base model.  
Unlike the large semantic gap between encoder output and vanilla image  in MAE, the base model target has similar semantics to the new model predictions. So a shallow 2-layer decoder is enough and better than the 8-layer decoder used in MAE. This design also greatly reduces the training cost.

\vspace{-5pt}
\section{Experiments}
\vspace{-10pt}
We evaluate our TEC  on ImageNet-1k~\citep{deng2009imagenet} by pretraining   randomly initialized  ViT~\citep{dosovitskiy2020image} with a 16$\times$16 patch size and 224$\times$224 image resolution for 300/800 epochs via AdamW~\citep{loshchilov2017decoupled}  of 4,096 batchsize. %~\citep{he2022masked}. 
To ensure the improvement is from TEC, we do not use any explicit/implicit extra training data and  the base models that are stronger than new model. 
Indeed, we respectively use   iBOT~\citep{zhou2021ibot} and MAE~\citep{he2022masked} pretrained ViT model  on ImageNet-1k  as our base model. 
We use the same masking strategy in MAE, \eg  75\% random masked ratio. See more training details in Appendix.

\begin{table}[t]
	\centering
	\setlength{\tabcolsep}{2mm}
	\renewcommand{\arraystretch}{0.8}
	\footnotesize
	\caption{Comparison with existing SSL methods under ImageNet-1k   finetuning using ViT. \dag~and gray color mean the usage of implicit /explicit extra data. 
	}
	\begin{tabular}{llccccccc}
		\toprule
         Model& Method          & Epoch  & Guidance & Top1 acc. \\	\midrule
         \multirow{28}{*}{ViT-Base}
         &DINO~\citep{caron2021emerging}  &  300 & NA & 82.8 \\
         &MoCov3~\citep{chen2021mocov3}   & 300 & NA & 83.2  \\
         &MixMIM~\citep{liu2022mixmim}   & 300 & RGB & 83.2 \\
         &MFM~\citep{xie2022masked}     & 300 & Frequency & 83.1 \\
         &\textcolor{gray}{BEiT~\citep{bao2021beit}}     & \textcolor{gray}{800} & \textcolor{gray}{DALLE\dag} & \textcolor{gray}{83.2} \\   
         &SplitMask~\citep{el2021large}  & 300 & NA & 83.6 \\
         &ConMIM~\citep{yi2022masked}   & 800 & Momentum & 83.7 \\
         &SimMIM~\citep{xie2022simmim}   & 800 & RGB & 83.8 \\
         &SIM~\citep{tao2022siamese}  & 1600 & Momentum & 83.8 \\
         &\textcolor{gray}{CAE~\citep{chen2022context}}      & \textcolor{gray}{1600}  & \textcolor{gray}{DALLE\dag} & \textcolor{gray}{83.9} \\
         &MaskFeat~\citep{wei2022masked}  & 1600 & HOG & 84.0 \\
         &LoMaR~\citep{chen2022efficient}    & 1600 & RGB & 84.1 \\
	     &BootMAE~\citep{dong2022ict} & 800 & RGB+Momentum & 84.2 \\  
         &data2vec~\citep{baevski2022data2vec}   & 800 & Momentum & 84.2 \\
         &Mugs~\citep{mugs2022SSL}  & 1600 & NA & 84.3 \\ 	
		 &\textcolor{gray}{MVP~\citep{wei2022mvp}}   & \textcolor{gray}{300} & \textcolor{gray}{CLIP\dag} & \textcolor{gray}{84.4} \\
         &PeCo~\citep{dong2021peco}     & 800 & Perceptual codebook & 84.5 \\
		 &CMAE~\citep{huang2022contrastive}  & 1600 & RGB & 84.7 \\
         &Ge2-AE~\citep{liu2022devil}   & 800 & RGB+Frequency & 84.8 \\
         &\textcolor{gray}{FD-CLIP~\citep{wei2022contrastive}}   & \textcolor{gray}{300} & \textcolor{gray}{CLIP\dag} & \textcolor{gray}{84.9} \\ \cline{2-5}
		 &MAE~\citep{he2022masked}      & 1600 & RGB & 83.6 \\
         &FD-MAE~\citep{wei2022contrastive}   & 300 & MAE & 83.8$_{+0.2}$ \\
         &\textbf{TEC}   & 300 & MAE  & \textbf{84.7$_{+1.1}$}  \\
         &\textbf{TEC}   & 800 & MAE  & \textbf{84.8$_{+1.2}$}  \\ \cline{2-5}
         	
         &\textcolor{gray}{iBOT-ImageNet-22K}   & - & \textcolor{gray}{Momentum} & \textcolor{gray}{84.4} \\ 
         &iBOT~\citep{zhou2021ibot}     & 1600 & Momentum & 84.1 \\ 
		 &SemMAE~\citep{li2022semmae} & 800 & iBOT & 84.5$_{+0.4}$ \\
         &\textbf{TEC}   & 300 & iBOT & \textbf{84.8$_{+0.7}$}  \\
         &\textbf{TEC}   & 800 & iBOT & \textbf{85.1$_{+1.0}$}  \\ \midrule
		 \multirow{2}{*}{ViT-Large} 
		 &MAE~\citep{he2022masked}      & 1600 & RGB & 85.9 \\
         &\textbf{TEC}   & 300 & MAE  & \textbf{86.5$_{+0.6}$}  \\
		\bottomrule
	\vspace{-20pt}
	\end{tabular}
	\label{tab:ftbase}
\end{table}

\begin{table}[tbp]
	\begin{minipage}[c]{.49\linewidth}
	\caption{Semantic segmentation on ADE20k using Upernet and ViT-B. }
		\label{tab:ade20k}
		\centering
		\setlength{\tabcolsep}{6mm} 
		\renewcommand{\arraystretch}{2.3}
		{ \fontsize{8pt}{3}\selectfont{
			\begin{tabular}{lcccc}
				\toprule
				Method &  Epoch & mIoU \\	\midrule
				BEiT & 800    &  47.1 \\
				PeCo &  800      & 48.5  \\
				GE2-AE & 800 & 48.9  \\
				CAE  & 1600     & 50.2 \\
				CMAE &  1600     & 50.1  \\ \midrule
				MAE	 &  1600      & 48.1   \\
				\textbf{TEC$_{\rm MAE}$}  & 800  & \textbf{49.9}   \\ \midrule
				iBOT	          & 1600 & 50.0   \\ 
				\textbf{TEC$_{\rm iBOT}$}  &  800 & \textbf{51.0}    \\
				\bottomrule
			\end{tabular}
		}}
	\end{minipage}
	\hfill
	\begin{minipage}[c]{.48\linewidth}
   \caption{Instance segmentation on COCO using Cascade MaskRCNN and ViT-B.
}
\label{tab:coco}
		\centering
		\setlength{\tabcolsep}{3mm} % column spacing
		\renewcommand{\arraystretch}{3.9}%  row spacing
	{ \fontsize{8pt}{3}\selectfont{
		\begin{tabular}{lcccc}
			\toprule
			Method         & AP$_{\rm bbox}$  & AP$_{\rm mask}$ \\	\midrule
			\multicolumn{3}{l}{Implementation from~\citep{zhou2021ibot}} \\ 
			iBOT	              & 51.2 & 44.2     \\
			\textbf{TEC$_{\rm iBOT}$} &  \textbf{52.7} & \textbf{45.4}     \\ \midrule
			\multicolumn{3}{l}{Implementation from~\citep{li2022exploring}} \\
			MAE	              & 54.0 &	46.7	     \\
			\textbf{TEC$_{\rm MAE}$} & \gsh{\textbf{54.6}}   & \gsh{\textbf{47.2}}           \\ 
			\bottomrule
		\end{tabular}
}}
	\end{minipage}%
	\vspace{-10pt}
\end{table}

\begin{table}[t]
	\begin{minipage}[c]{.49\linewidth}
	\caption{Top1 accuracy on the ImageNet-1k dataset under 
	parameter-efficient finetuning.}
		%		\vspace{0.1em}
		\label{tab:prompting}
		\centering
		\setlength{\tabcolsep}{4.2pt} % column spacing
		\renewcommand{\arraystretch}{3.}
		%\footnotesize{
		{ \fontsize{8pt}{3}\selectfont{
			\begin{tabular}{lccc}
				\toprule
				Method	& Epoch	& Settings   & Top 1 acc.    \\	\midrule
				MAE     & 1600  & Linear probing & 68.0   \\ \midrule
				\multirow{3}{*}{TEC$_{\rm MAE}$} & \multirow{3}{*}{800} & Linear probing & 69.8   \\ 
				& & +Input adapter FT   & 72.6 \\
				& & +Encoder adapter FT & 79.9     \\
				\bottomrule
			\end{tabular}
		}}
	\end{minipage}
	\hspace{0.25cm}
	\begin{minipage}[c]{.48\linewidth}
   \caption{Semi-supervised semantic segmentation on the ImageNet-S dataset.
}
\label{tab:imagenets}
		\centering
		\setlength{\tabcolsep}{3.8pt} % column spacing
		\renewcommand{\arraystretch}{3.}%  row spacing
	{ \fontsize{8pt}{3}\selectfont{
		\begin{tabular}{lcccc}
			\toprule
			Pretrain		& Method	& Epoch       & mIoU$_{\rm val}$  \\	\midrule
			\multirow{2}{*}{SSL} & MAE	        & 1600        & 38.3    \\
			&TEC$_{\rm MAE}$  & 800    & \textbf{42.9}   &            \\ \midrule
			\multirow{2}{*}{SSL+FT}&MAE	        & 1600+100        & 61.0        \\  
			&TEC$_{\rm MAE}$  & 800+100    & \textbf{62.0}     &            \\ 
			\bottomrule
		\end{tabular}
}}
	\end{minipage}%
	\vspace{-10pt}
\end{table}

\subsection{Performance comparison}
\vspace{-5pt}
\subsubsection{Comparison on  ImageNet }
\vspace{-5pt}
\myPara{Finetuning on ImageNet-1k.} \tabref{tab:ftbase} summarizes the   finetuning performance on ImageNet-1k. 
One can observe  that with iBOT as base model, TEC surpasses the base model by 0.7\% under   300 training epochs from random initialization, and further makes  1.0\% improvement for 800 epochs.  
Similarly, TEC respectively brings 1.1\% and 1.2\% improvement over the MAE base model under  300/800 training epochs. 
These results show that  thanks to the proposed target-enhanced conditional MIM scheme, TEC actually can further improve the strong MIM-based methods, \eg~MAE and iBOT used here. %  verifying its effectiveness. 
Besides, \tabref{tab:ftbase}  also shows that under similar or even less training cost,  TEC outperforms other SOTA SSLs, including methods trained by implicitly extra data, \eg MVP~\citep{wei2022mvp} and FD-CLIP~\citep{wei2022contrastive}.  

More surprisingly,  TEC with only ImageNet-1k data has an improvement of 0.7\% over iBOT trained on  ImageNet-22k, indicating more effectiveness of TEC pretraining over more training data.   
To the best of our knowledge, \textit{TEC sets a new SOTA 85.1\% on ViT-B when solely using ImageNet-1k}, showing the potential of sustainable SSL learning. 
We also investigate the scaling ability of TEC by using ViT-Large, and observe that TEC surpasses the MAE pretrained base model by 0.6\% with 300 epochs from random initialization.

\myPara{Parameter-efficient finetuning on ImageNet-1k.}
Parameter-efficient finetuning methods, \eg linear probing, aims to  finetune a small amount  of parameters for adapting to a downstream task. 
We test TEC under the linear probing setting which  only finetunes a linear classifier at the top of a frozen pretrained model. \tabref{tab:prompting} reports the classification accuracy  of  ViT-B on ImageNet-1k under linear probing.  TEC improves the MAE base model by 1.8\%, showing more category-related semantic information in the learned new model. 
Indeed, our input and encoder adapters used for pretraining can also be used for parameter-efficient finetuning. 
Finetuning input adapter via prompting  brings a remarkable improvement  of 4.6\%, and finetuning both  input and encoder adapters makes  11.9\% improvement over the MAE base model. This also shows the benefit of our proposed adapters. 

\myPara{Semantic segmentation on ImageNet-S.} 
To test the pixel-level representation ability of TEC pretrained models, we conduct semantic segmentation finetuning on ImageNet-S~\citep{gao2021luss} which has  pixel-level training labels on ImageNet.
We use ViT-B as the segmentation model without   extra segmentation head, since the pretraining and finetuning data have no domain shift. 
 \tabref{tab:imagenets} shows that 
 TEC$_{\rm MAE}$ improves   MAE base model by 4.6\% on  mIoU. When using  supervised ImageNet fully-finetuned pretraining, TEC$_{\rm MAE}$ achieves a gain of 1.0\% over MAE.

\vspace{-10pt}
\subsubsection{Transfer learning on downstream tasks}
\vspace{-10pt}
Here we investigate the transfer learning ability of TEC models on downstream tasks.

\myPara{Semantic segmentation.}
For semantic segmentation on the ADE20k~\citep{zhou2018semantic} dataset, we use Upernet~\citep{xiao2018unified} with  ViT-B as the segmentation model.
\tabref{tab:ade20k} shows that TEC$_{\rm iBOT}$ surpasses the iBOT base model by 1.0\%   mIoU, and
TEC$_{\rm MAE}$ achieves a 1.8\% improvement  over its MAE base model.
Thus, TEC pretrained models show greater transfer learning abilities on semantic segmentation compared to their base models.
Also, TEC shows remarkable advantages over strong competitors with fewer pretraining epochs. For example, it outperforms MAE, CAE~\citep{chen2022context}, and CMAE~\citep{huang2022contrastive} by 2.9\%, 0.8\%, and 0.9\%, achieving new SOTA.

\myPara{Instance segmentation.}
For instance segmentation on COCO~\citep{lin2014microsoft}, for fairness, we apply the Cascade MaskRCNN~\citep{cai2019cascade} implemented by iBOT~\citep{zhou2021ibot} and ViTDet~\citep{li2022exploring} for TEC with iBOT/MAE base models.
 \tabref{tab:coco} shows that with the  implementation from iBOT, TEC surpasses the iBOT base model by 1.5\% on box AP and 1.2\% on mask AP; and when using the implementation from ViTDet, TEC also achieves a gain of 0.6\% on box AP and 0.5\% on mask AP, indicating stable improvements of TEC.
 \vspace{-10pt}
\subsection{Ablation and Analysis}
\begin{wrapfigure}{r}{5cm}
 \vspace{-20pt}
\centering
\tiny
\begin{overpic}[width=0.9\linewidth]{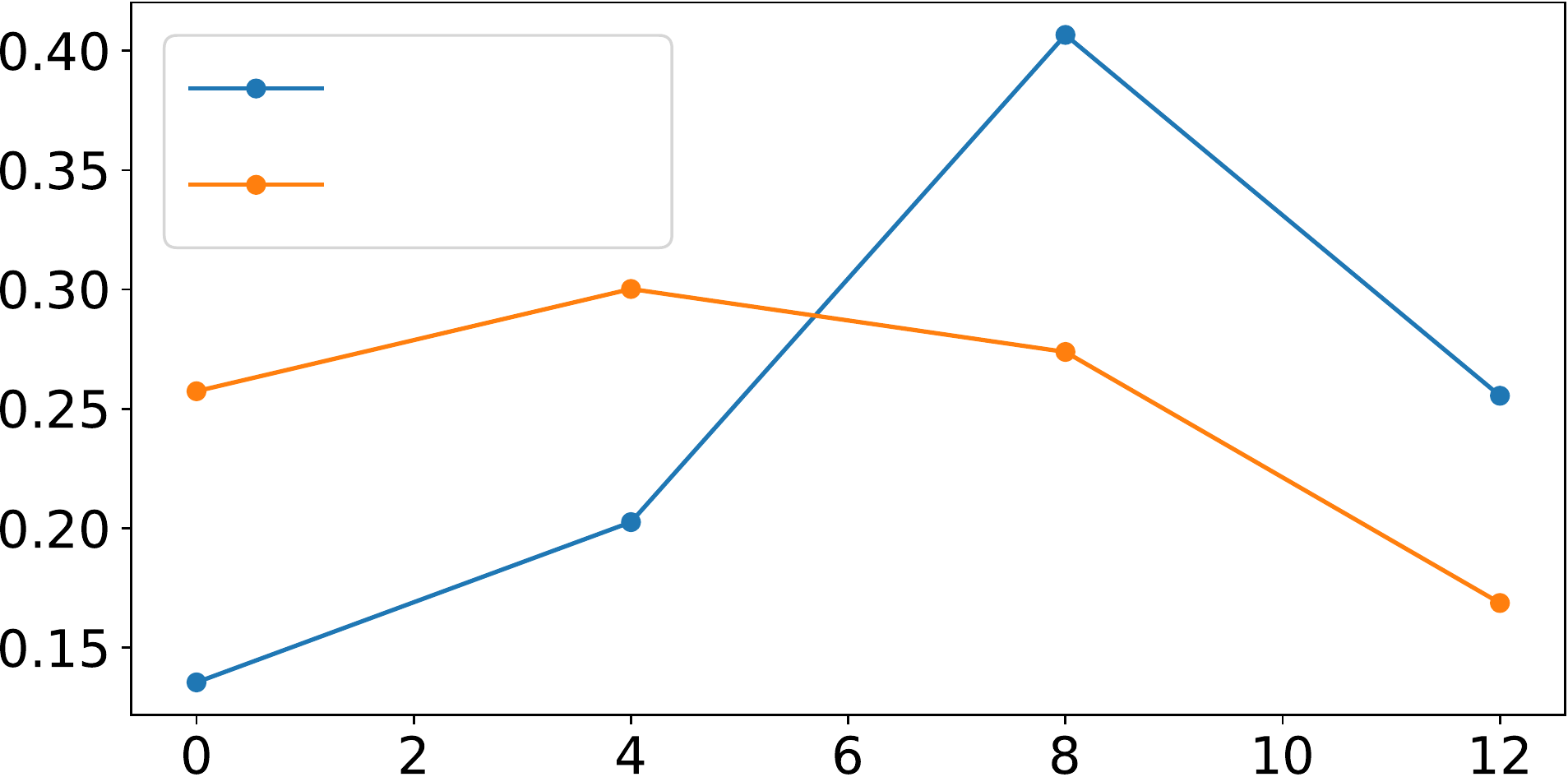} %
	\put(-5,10){\rotatebox{90}{Average Proportion}}
	\put(30,-3){ViT Encoder Layers}
	\put(21,43){TEC$_{\rm iBOT}$}
	\put(22,37){TEC$_{\rm MAE}$}
\end{overpic}
    \vspace{-5pt}
   \caption{The average proportion of encoder adapters contributing
   to the encoder output $Z_{e}$.}
  \label{tab:adap_compare}
	\vspace{-10pt}
\end{wrapfigure}
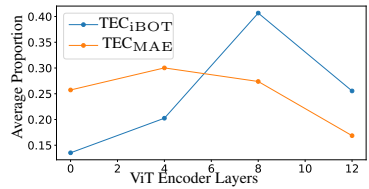
We give the ablation study and analysis of our TEC.
By default, models are pretrained with 300 epochs and evaluated with the  finetuning on ImageNet-1k. 

\myPara{Conditional pretraining.}
The conditional adapters aid the SSL pretraining under different base models.
\tabref{tab:abl_all} shows adapters stably improve the performance by 0.4\% and 0.2\% when using MAE and iBOT as base models.
To observe the adaptation difference to base models, we show the average proportion of encoder adapters contributing to the encoder output in~\figref{tab:adap_compare},
\ie $Z_{n}^{'}/Z_{e}$ in Eqn.~\ref{targets}.
 iBOT base model requires adapters to provide more features from deeper layers, while  MAE base model makes adapters focus more on shallow layers,
which is constant with their properties, \ie iBOT base model has more high-level category semantics while MAE model has more low-level image details.

\myPara{Feature normalization on different dimensions.}
We normalize target features on the patch dimension to stress the relative relations among patches,
which differs from existing methods that normalize
features on the channel dimension.
In~\tabref{tab:norm_dim}, normalizing on patch dimension achieves a 0.3\% gain than channel-dim normalization.
In contrast, the channel-dim normalization has no effect compared to the unnormalized version.
Channel-dim normalization emphasizes the feature difference in channels.
Instead, our patch-dim normalization stresses the relations among patches, which is compatible with the patch prediction in the MIM scheme.
\tabref{tab:abl_all} shows training with patch-dim normalized feature has the 0.6\%/0.4\% gain over MAE/iBOT base models, showing its robustness over base models.

\myPara{Semantic-related attention.}
The KQ attention maps naturally contain the semantic relations among patches
and thus are used as the base model targets with enhanced patch-relation properties.
 \tabref{tab:abl_all} shows that using attention maps further improves the models trained with patch-dim normalization.
\tabref{tab:attention_type} compares the effects of different types of attention maps.
Only using the attention maps of the class token has no improvement, while the attention of semantic-related patches brings a 0.2\% gain over the baseline.
Therefore, it is the relation among patches that helps the MIM training. 
Compared to using all attention maps,
using the selected semantic-related attention maps
brings a larger gain by reducing the noise.

\begin{table}
	\caption{Ablation study on ImageNet-1K fully finetuning setting using ViT-B.}
	\centering
	\footnotesize
	\subtable[Ablation study of proposed modules.]{
		\centering
		\setlength{\tabcolsep}{4mm}
		\renewcommand{\arraystretch}{0.6}
	\footnotesize
	\begin{tabular}{lcccccccc}
		\toprule
         Patch-norm. feature      & Semantic attention & Adapters & MAE base & iBOT base \\	\midrule
		 \multicolumn{3}{c}{Base model performance}    & 83.6\%  & 84.1\% \\ \midrule
         \checkmark   &           &          & 84.2\%  & 84.5\%  \\ 
         \checkmark   & \checkmark&          & 84.3\% & 84.7\% \\
		 \checkmark   &           & \checkmark& 84.6\%   &  84.7\% \\
		 \checkmark   & \checkmark& \checkmark& 84.7\%   & 84.8\%\\
		\bottomrule
	\end{tabular}
	\label{tab:abl_all}
	}
	\subtable[Effect of adapters.]{
		\centering
		\setlength{\tabcolsep}{2.0mm}
	\renewcommand{\arraystretch}{0.6}
		\begin{tabular}{lccccccc}
			\toprule
						               & Top1 acc.  \\	\midrule
			 MAE base &                    83.6  \\ \hline
			 No adapter            & 84.2     \\ 
 			 + input adapter       & 84.3     \\
			 + encoder adapter     & 84.6            \\
			\bottomrule
		\end{tabular}
		\label{tab:adapters}
	}
	\subtable[Patch-norm features.]{
		\centering
		\setlength{\tabcolsep}{2.0mm}
	\renewcommand{\arraystretch}{0.6}
		\begin{tabular}{lcccc}
			\toprule
					 & Top1 acc. \\	\midrule
			 MAE base &                    83.6  \\ \hline
			 NA                        &  83.9  \\
			 Feature dim.              &  83.9      \\ 
			 \textbf{Patch dim.}       &  \textbf{84.2}      \\
			\bottomrule
		\end{tabular}
		\label{tab:norm_dim}
	}
	\vspace{-5pt}
	\subtable[Init. with base model pretrain.]{
		\centering
		\renewcommand{\arraystretch}{1.}
		\setlength{\tabcolsep}{4mm}
		\begin{tabular}{lcccccccc}
			\toprule
					 & Top1 acc. \\	\midrule
			iBOT base & 84.1 \\ \hline
			Load     & 84.4           \\
			\textbf{Not load}  & \textbf{84.8}      \\ 
			\bottomrule
		\end{tabular}
		\label{tab:loadpretrain}
	}
	\subtable[TEC accelerates MAE training.]{
		\centering
		\setlength{\tabcolsep}{4mm}
		\renewcommand{\arraystretch}{0.6}
		\begin{tabular}{lcccccccc}
			\toprule
			 & Epoch & Top1 acc.  \\	\midrule
			MAE & 1600 & 83.6 \\ 
			\textbf{TEC$_{\rm MAE1600ep}$} & 300 & \textbf{84.7$_{+1.1}$}  \\ \hline
			MAE & 300  & 82.9           \\ 
			\textbf{TEC$_{\rm MAE300ep}$} &\textbf{100}  & \textbf{83.9$_{+1.0}$}   \\ 
			\textbf{TEC$_{\rm MAE300ep}$} &300   & \textbf{84.3$_{+1.4}$}   \\ 
			\bottomrule
		\end{tabular}
		\label{tab:mae300ep}
	}
	\subtable[Effect of semantic-related patch attention.]{
		\centering
		\setlength{\tabcolsep}{6mm}
		\renewcommand{\arraystretch}{0.6}
		\begin{tabular}{lcccccccc}
			\toprule
			 		  & Top1 acc. \\	\midrule
			 iBOT base    & 84.1 \\ \hline
			 No attention     & 84.5      \\ 
			 Cls token only  & 84.5  \\ 
			 All attention     & 84.6   \\
			 \textbf{Attention select}   & \textbf{84.7}  \\
			\bottomrule
		\end{tabular}
		\label{tab:attention_type}
	}
	\vspace{-25pt}
\end{table}

\myPara{Accelerating the training process of base models.}
By default, we use the fully pretrained SSL models as base models.
To verify if TEC can improve an unconverged SSL model, we use a 300-epoch MAE pretrained ViT-B as the base model and train TEC with 100/300 epoch from random initialization.
As shown in~\tabref{tab:mae300ep}, the 300-epoch pretrained MAE gives a 82.9\% Top.1 accuracy.
In comparison, TEC$_{\rm MAE300ep}$ achieves 84.3\%/83.9\% with 300/100 epochs, surpassing the 300-epoch MAE base models with 1.4\%/1.0\%.
Notably, TEC$_{\rm MAE300ep}$ even outperforms the 1600-epoch pretrained MAE by 0.3\% with only 100 epoch training, showing TEC can significantly accelerate the training process of the base model.
\begin{wraptable}{r}{4.5cm}
	\vspace{-20pt}
	\caption{Towards general sustainable SSL 
	using the TEC as the new base model.}
		\label{tab:again}
		\centering
		\setlength{\tabcolsep}{0.3mm} % column spacing
		 \renewcommand{\arraystretch}{2.5}
		{ \fontsize{8pt}{3}\selectfont{
			\begin{tabular}{lcccc}
				\toprule
				Model	 &  Base & Epoch & Top1 acc. \\	\midrule
				iBOT &  -   &  1600 & 84.1 \\
				TEC$_{\rm iBOT}$ &  iBOT   &  800 & 85.1 \\
				TEC &  TEC$_{\rm iBOT}$   &  800 & 85.2 \\
				\bottomrule
			\end{tabular}
		}}
	\vspace{-20pt}
	\end{wraptable}
Still, the TEC$_{\rm MAE1600ep}$ taught with 1600-epoch MAE base model further improves the TEC$_{\rm MAE300ep}$ by $0.4\%$,
indicating our sustainable learning scheme relies on good base models to achieve better performance.

\myPara{Towards general sustainable SSL.}
To make more steps towards sustainable SSL,  we use the TEC pretrained models as the base model for a new round of TEC pretraining.
\tabref{tab:again} shows that the second-round TEC trained with the first-round TEC base model  achieves 85.2\%.
The possible reason for the smaller improvement in the second round is caused by the limited network capacity or 
two rounds of TEC pretraining learns similar knowledge.

\myPara{Initializing new model with base model weights or not.}
The new models in the TEC framework are trained from random initialization.
\tabref{tab:loadpretrain} compares the new model performance with/without loading the pretrained weights of the base model.
The randomly initialized new model outperforms the model loaded with pretrained base model weights by 0.4\%.
We assume that randomly initialized new models avoid the local minima of the base model,
and the new model learns a different weight distribution from the base model.

\section{Related works}
\vspace{-2pt}
\label{sec:tab}
\myPara{Self-supervised learning.}
Self-supervised learning enables representation pretraining without human annotation
by training with pretext tasks, \eg instance discrimination task (ID)
and masked image modeling task (MIM).
ID learn high category-related representations by pulling close representations from multiple views of one image~\cite{chen2020simple,byol,Chen_2021_CVPR,barlow_Twins,caron2020unsupervised}.
Extracting representations from multi-view requires larger training cost than supervised training.
MIM learns semantics by reconstructing the information of masked patches from unmasked parts,
which learns more spatial semantic details than ID.
Due to the incomplete input, MIM
usually requires longer training epochs than ID to converge.
\citep{huang2022contrastive,wang2022repre} explore combining the advantages of MIM and ID to further improve performance.
Recently, \citep{kong2022understanding} reveals both MIM and ID learn occlusion-invariant features.
We observe a trend that these SSL methods require increasingly large computing costs to achieve SOTA, which hinders the development of new SSL methods.
To remedy this, we explore sustainable SSL by learning from pretrained SSL models.

\myPara{Masked image modeling on various targets.}
The reconstruction targets guide the MIM learning on different semantic spaces. 
MIM has explored various reconstruction targets, \eg RGB pixels and tokenizers.
To make images similar to the discretized language in NLP~\citep{devlin2018bert},  
Beit~\citep{bao2021beit}
utilizes the DALLE pretrained tokenizer~\citep{ramesh2021zero}
as the prediction target.
CAE~\citep{chen2022context} further decouples this pretext task prediction with the encoder. 
MAE and SimMIM~\citep{xie2022simmim} show using RGB images as targets achieves competitive fully finetuning performance.
MaskFeat~\citep{wei2022masked} reveals hand-designed HOG feature~\citep{dalal2005histograms} is an effective target form.
Ge2-AE~\citep{liu2022devil} and MFM~\citep{xie2022masked} find the image frequency domain can be complementary to RGB image targets.
The perceptual codebook in PeCo~\citep{dong2021peco} helps the model learn semantic information.
The online momentum network~\citep{He_2020_CVPR}
is used by iBOT and data2vec~\citep{baevski2022data2vec}
to provide updated prediction targets.
BootMAE~\citep{dong2022ict} takes advantage of RGB images and online network
targets.
\citep{yang2022masked} enhances the distillation from a large teacher model to a compact student
model with the masking scheme.
MVP~\citep{wei2022mvp} introduces rich semantics learned from vision-language pretraining
using the CLIP pretrained model~\citep{ramesh2021zero} as the target.
Unlike these works that stress the unique properties of a specific reconstruction target,
we show that all SSL pretrained models can serve as good base models with the help of target-enhancement in TEC.
The adapters and target-enhancing scheme
in TEC enables the good adaptability to various base model targets.

\myPara{Self-supervised knowledge distillation.}
The self-supervised knowledge distillation is similar to sustainable learning in this work
as they both learn from SSL pretrained models.
However, knowledge distillation methods mostly aim to transfer knowledge from strong and large models to compact new models.
SEED \citep{fang2021seed} distillates knowledge from large SSL models to small models with contrastive loss.
\citep{navaneet2022simreg} uses MLP heads for feature regression to distill large SSL teachers to compact student models.
\citep{xu2021bag} groups instances with teacher models and transfers the instance-relation knowledge to student models.
As an exception, \citep{wei2022contrastive} shows feature distillation improves contrastive-based SSL models but brings marginal gain over the SOTA MAE~\citep{he2022masked} model.
In contrast, our method for sustainable learning lets the new model surpass the base model with the same model size and training data.

\vspace{-3pt}
\section{Conclusion}
\vspace{-3pt}
This work explores sustainable self-supervised learning by learning from pretrained SSL models.
We propose a target-enhanced conditional mask-reconstruction learning scheme to learn from and surpass existing SSL models.
The adapters help to adapt the new model to various base models during pretraining
and can also serve as parameter-efficient finetuning modules.
We utilize the mask-reconstruction scheme as the basis for surpassing base models,
and we construct prediction targets with enhanced patch-level relations to aid the MIM pretraining.
Our method further improves the strong MIM pretrained methods, \eg MAE and iBOT, 
proving the feasibility of sustainable learning.

\section{Acknowledgements}
We thank for the helpful discussion with Quanhong Fu, Xingyu Xie and Chengze Lu.

\bibliography{ref}
\bibliographystyle{iclr2023_conference}

\clearpage
\appendix
\section{Appendix}
\paragraph{Pretraining settings on ImageNet-1k.}
We use the standard ViT network implemented in MAE.
We give the pretraining settings in~\tabref{tab:pretrain_weights},
which follows the pretraining settings in MAE.
Due to the different properties of SSL base models,
we set different parameters of semantic-related attention for MAE and iBOT base models, as shown in~\tabref{tab:sem_att_weights}.

Inspired by~\citep{baevski2022data2vec}, we utilize the average output features of last two blocks of the base model as the feature target.
We measure the CKA similarity~\citep{kornblith2019similarity} of each mid-layer block to the output of the last block,
and we observe the high feature similarity of last two blocks.

\paragraph{Fully finetuning settings on ImageNet-1k.}
We give the fully finetuning settings on ImageNet-1k in~\tabref{tab:finetune}.
We observe that TECs trained with different base models may have different properties.
Since these base models have different finetuning layer decay values,
we set different layer decay values for TECs trained with different base models.

\paragraph{Parameter-efficient finetuning settings on ImageNet-1k.}
Following MAE, the linear probing settings are shown in~\tabref{tab:liner_prob}.
For parameter-efficient finetuning with the input adapter,
we use the same training settings as used by the liner probing in~\tabref{tab:liner_prob}.
When finetuning with the encoder adapters,
we use the same training settings as used by the fully finetuning in~\tabref{tab:finetune} due to more parameters are contained in encoder adapters.

\paragraph{Semi-supervised semantic segmentation finetuning on ImageNet-S.}
We give the training settings of semi-supervised semantic segmentation finetuning on ImageNet-S in~\tabref{tab:imagenet-seg}.
We set different learning rates and layer decay weights for models
initialized with pretrained weights with/without fully finetuning.

\paragraph{Downstream task settings.}
For semantic segmentation on ADE20K, 
we use the MMSegmentation~\citep{mmseg2020} implementation of Upernet.
The training configurations follow the MAE training configurations in MMSegmentation.
For instance segmentation on COCO,
we follow the training configurations of iBOT and ViTDet
for TEC$_{\rm iBOT}$ and TEC$_{\rm MAE}$.

\begin{table}[h]
	\centering
	\setlength{\tabcolsep}{8mm}
	\caption{Pretraining settings.
	}
	\small
	\begin{tabular}{lcccccccc}
		\toprule
        Configuration      & Value \\	\midrule
        Optimizer & AdamW  \\
        Base learning rate & 1.5e-4 \\
        Weight decay & 0.05 \\
        Optimizer momentum & $\beta_1, \beta_2{=}0.9, 0.95$  \\
        Batch size & 4096 \\
        Learning rate schedule & Cosine decay\\
        Warmup epochs  & 40 \\
        Augmentation & RandomResizedCrop \\
		\bottomrule
	\end{tabular}
	\label{tab:pretrain_weights}
\end{table}

\begin{table}[]
	\centering
	\setlength{\tabcolsep}{10mm}
	\caption{Parameters of semantic-related attention.
	}
	\small
	\begin{tabular}{lcccccccc}
		\toprule
        Base model       &$\tau$ & k \\	\midrule
        MAE (ViT-Base) & 1.8 & 15  \\
        MAE (ViT-Large) & 1.4 & 15  \\
        iBOT (ViT-Base) & 1.0 & 9  \\
		\bottomrule
	\end{tabular}
	\label{tab:sem_att_weights}
	\vspace{-15pt}
\end{table}
\begin{table}[]
	\centering
	\setlength{\tabcolsep}{8mm}
	\caption{Settings of fully finetuning and parameter-efficient finetuning with encoder adapters.
	}
	\small
	\begin{tabular}{lcccccccc}
		\toprule
        Configuration      & Value \\	\midrule
        Optimizer & AdamW \\
        Base learning rate & 1e-3 \\
        Min learning rate & 1e-6 (B), 1e-5(L)  \\
        Weight decay & 0.05 \\
        Optimizer momentum & $\beta_1, \beta_2{=}0.9, 0.999$ \\
        Layer-wise lr decay  & 0.55 (MAE-B), 0.65 (iBOT-B), 0.65 (MAE-L)  \\
        Batch size & 1024 \\
        Learning rate schedule & Cosine decay \\
        Warmup epochs & 20 (B), 5 (L) \\
        Training epochs & 100 (B), 50 (L) \\
        Augmentation & RandAug (9, 0.5)  \\
        Label smoothing & 0.1 \\
        Mixup  & 0.8 \\
        Cutmix  & 1.0 \\
        Drop path  & 0.1 \\
		\bottomrule
	\end{tabular}
	\label{tab:finetune}
\end{table}

\begin{table}[]
	\centering
	\setlength{\tabcolsep}{8mm}
	\caption{Settings of linear probing and parameter-efficient finetuning with the input adapter.
	}
	\small
	\begin{tabular}{lcccccccc}
		\toprule
        Configuration      & Value \\	\midrule
        Optimizer & LARS \\
        Base learning rate & 0.1 \\
        Weight decay & 0 \\
        Optimizer momentum & 0.9 \\
        Batch size & 16384 \\
        Learning rate schedule & Cosine decay \\
        Warmup epochs & 10 \\
        Training epochs & 90 \\
        Augmentation & RandomResizedCrop \\
		\bottomrule
	\end{tabular}
	\label{tab:liner_prob}
\end{table}

\begin{table}[]
	\centering
	\setlength{\tabcolsep}{8mm}
	\caption{Settings of semantic segmentation finetuning on ImageNet-S.
	}
	\small
	\begin{tabular}{lcccccccc}
		\toprule
        Configuration      & Value \\	\midrule
        Optimizer & AdamW \\
        Base learning rate & 5e-4 (SSL), 1e-4 (SSL+FT) \\
        Weight decay & 0.05 \\
        Optimizer momentum & $\beta_1, \beta_2{=}0.9, 0.999$ \\
        Layer-wise lr decay  & 0.60 (SSL), 0.45 (SSL+FT)  \\
        Batch size & 256 \\
        Learning rate schedule & Cosine decay \\
        Warmup epochs & 5 \\
        Training epochs & 100  \\
        Augmentation & RandomResizedCrop \\
        Drop path  & 0.1 \\
		\bottomrule
	\end{tabular}
	\label{tab:imagenet-seg}
\end{table}

\end{document}